\documentclass[article]{jss}

\usepackage{thumbpdf,lmodern}

\usepackage{framed}
\usepackage{amsmath}
\usepackage{graphicx,psfrag,epsf}
\usepackage{enumerate}
\usepackage{algorithm}
\usepackage{algorithmicx}
\usepackage{algpseudocode}
\usepackage{xcolor}
\usepackage{natbib}
\usepackage{url} 
\usepackage{nameref}
\usepackage{listings}
\newcommand{\fct}[1]{\code{#1()}}
\newcommand{\factorMerger}{\pkg{factorMerger }}

\newcommand{\M}{\mathcal{M}}

\author{Agnieszka Sitko \\University of Warsaw
   \And Przemys\l{}aw Biecek\\Warsaw University of Technology }
\Plainauthor{Agnieszka Sitko, Przemyslaw Biecek}

\title{The Merging Path Plot: adaptive fusing of $k$-groups with likelihood-based model selection}
\Plaintitle{The Merging Path Plot: adaptive fusing of k-groups with likelihood-based model selection}
\Shorttitle{The Merging Path Plot: likelihood-based adaptive fusing of k-groups}

\Abstract{
There are many statistical tests that verify the null hypothesis: the variable of interest has the same distribution among $k$-groups. But once the null hypothesis is rejected, how to present the structure of dissimilarity between groups?

  In this article, we introduce The Merging Path Plot --- a~methodology, and \factorMerger --- an \proglang{R}\ package, for exploration and visualization of $k$-group dissimilarities. Comparison of $k$-groups is one of the most important issues in exploratory analyses and it has zillions of applications. The classical solution is to test a~null hypothesis that observations from all groups come from the same distribution. If the global null hypothesis is rejected, a~more detailed analysis of differences among pairs of groups is performed. The traditional approach is to use pairwise \emph{post hoc tests} in order to verify which groups differ significantly. However, this approach fails with a large number of groups in both interpretation and visualization layer. The~Merging Path Plot methodology solves this problem by using an easy-to-understand description of dissimilarity among groups based on Likelihood Ratio Test (LRT) statistic.
}

\Plainkeywords{Post-hoc testing, 
hierarchical clustering, likelihood~ratio test, R}

\Address{
  Agnieszka Sitko\\
  Faculty of Mathematics,  
    Informatics and Mechanics \\
    University of Warsaw \\
  E-mail: \email{ag.sitko@gmail.com} 
    \vspace{1em}
	\\
  Przemys\l{}aw Biecek\\
  Faculty of Mathematics and Information Science,\\
	Warsaw University of Technology \\
  E-mail: \email{przemyslaw.biecek@gmail.com}\\
  URL: \url{http://biecek.pl/}
}

\begin{document}



\section{Introduction and Motivation}
\label{sec:intro}
One of the most frequent tasks in exploratory analyses is comparison of $k$ groups. There are zillions of applications, such as comparisons of different medical treatments, comparisons of different countries or comparisons of segments of clients. The classical solution is to test the global hypothesis that all groups are equal. 
If the global null hypothesis is rejected, a more detailed analysis of differences among pairs of groups is needed. The traditional approach is to perform \emph{post hoc tests} in order to verify which groups differ significantly.

\begin{figure}[h!tbp]
\centering
\includegraphics[scale=0.6]{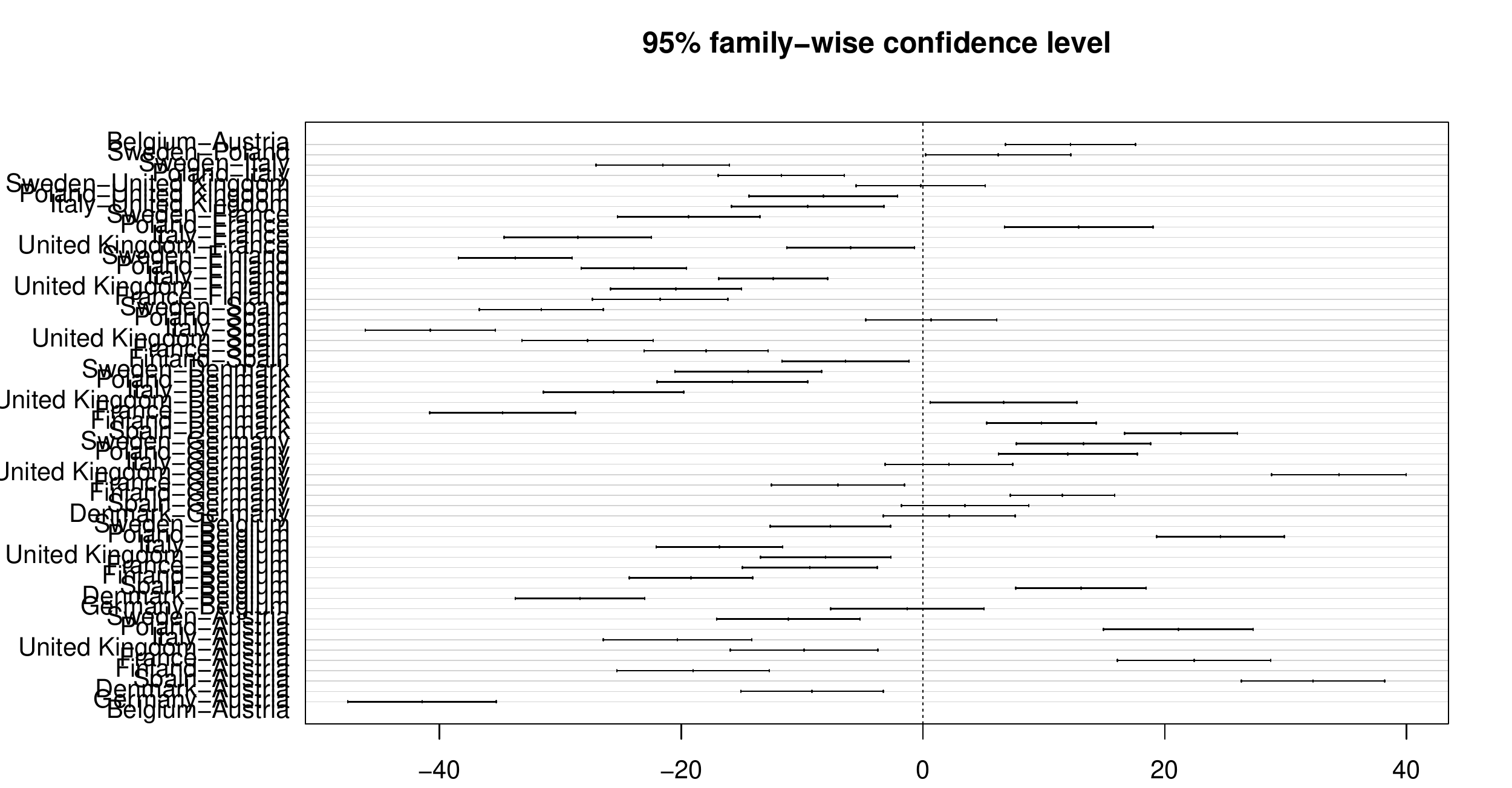}
\caption{\label{fig:postHocPISA} Classical approach to graphical presentation of post-hoc testing of 11 groups with the use of the \code{plot.tukeyHSD} function. 
This plot is based on the data from 11 counties (55 pairs) selected from PISA 2012 dataset.
For each pair of countries the plot presents average differences and 95\% confidence intervals.}
\vspace{5mm}

\includegraphics[scale=0.6]{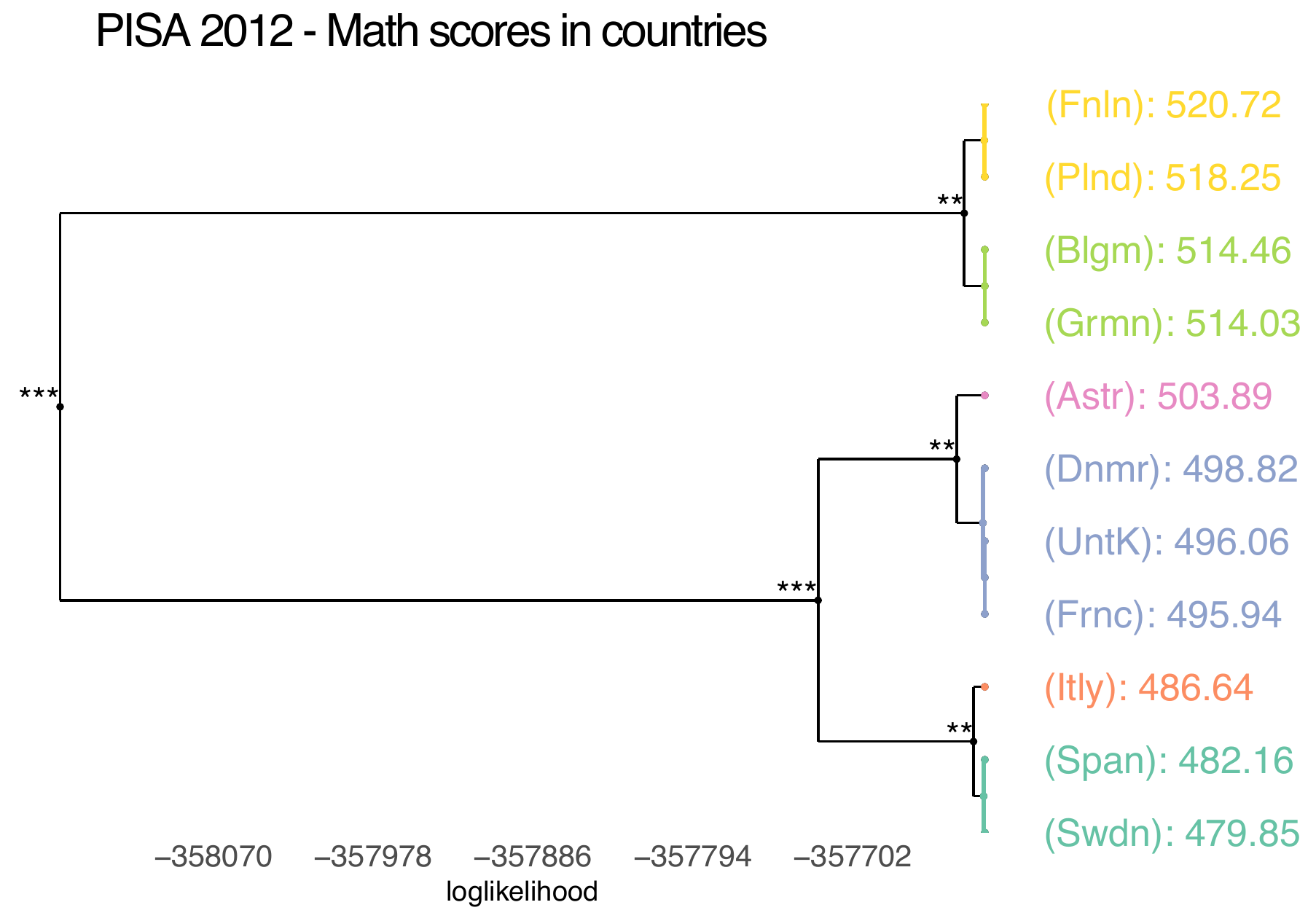}
\caption{\label{fig:FMtukeydone2}The Merging Path Plot for data from 11 counties selected from PISA 2012 (the same data as in Figure \ref{fig:postHocPISA}). 
Colors annotate groups of countries with similar averages. Numbers displayed on the right side of country names stand for country averages. The position on which branches are merged corresponds to the likelihood of a model with combined groups. Stars denote p-values from LRT test between consecutive models.}
\end{figure}

\clearpage

As we will show later, this approach fails if the number of groups is large as the number of pairs quickly grows beyond easy interpretation.

The larger the number of groups, the more pronounced is the problem with classical post-hoc testing. 
For example, in the \textit{Programme for International Student Assessment} (PISA) \citep{pisa2012} data about academic performance of 15-year-old kids from 65 countries is collected. One can use tests such as \textit{Analysis of variance} (ANOVA) or other $k$-sample tests to verify whether there are any differences between countries but then the question arises as to the nature of the identified differences between the countries. The total number of pairwise comparisons is $\frac{65(65-1)}{2}=2080$ and obviously it is not easy to present such a number of results in an easy-to-understand way. Figure \ref{fig:postHocPISA}, where results for only 11 European countries are presented, shows how hard it is to read anything when the number of groups is not small.

The problem with post-hoc testing is also related to the inconsistency of results. For a fixed significance level, it is possible that the mean in group A does not differ significantly from the one in group B, and similarly with groups B and C. At the same time, the difference between group A and C is detected. Then data partition is unequivocal and, as a consequence, impossible to put through. 

To deal with this problem, we introduce the Merging Path Plot methodology along with \pkg{factorMerger} --- a library for \proglang{R} \citep{Rcran}. The aim of the methodology is to enrich results from k-groups test and provide a greater variety of plots designed for deeper understanding of analyzed models. 
An example is presented in Figure \ref{fig:FMtukeydone2}.

The aim of the \factorMerger package is to provide informative and easy-to-understand visualizations of post-hoc comparisons. 
It works for a wide spectrum of probability distribution families. 
The Merging Path Plot shows consistent and non-overlapping adaptive fusing of groups based on the \textit{Likelihood Ratio Test} (LRT) statistics. In addition, the \textit{Generalized Information Criterion} (GIC) is presented for fused models. This criterion may be used to choose the optimal fusion of groups. 

\section{Background and Related Work}

One may find implementations of the traditional \emph{post-hoc tests} in many \proglang{R}~packages. For example, package \pkg{agricolae} \citep{Agric} offers a wide range of them. It gives one of the most popular \emph{post-hoc test}, Tukey HSD test (function \fct{HSD.test}), its less conservative version --- Student-Newman-Keuls test (function \fct{SNK.test}) or Scheffe test (function \fct{scheffe.test}), which is robust to group imbalance. These parametric tests are based on Student's t-distribution and are thus reduced to Gaussian models only. In contrast, \pkg{multcomp} package \citep{Multcomp} can be used with generalized linear models (function \fct{glht}) as it uses general linear hypothesis. Similarly to the \pkg{multcomp}, some implementations that accept \fct{glm} objects are also given in \pkg{car} (\fct{linearHypothesis}, \citealp{car}) and \pkg{lsmeans} \citep{lsmeans}.

But what about the problem of clustering categorical variables into non-overlapping groups? It has already been presented in the literature. 
The first person to propose an iterative procedure for merging factor levels based on the studentized range distribution was John Tukey \citep{Tukey}. However, again, the statistical test used in this approach rendered it limited to Gaussian models.

\clearpage

\emph{Collapse And Shrinkage in ANOVA} (\emph{CAS-ANOVA}, \citealp{Casanova}) is an algorithm that extends categorical variable partitioning for generalized linear models. It~is~based on the Tibshirani's \emph{Fused LASSO} \citep{Tib} with constraints imposed on pairwise differences within a factor, which yields their smoothing.
Yet another approach that is also adjusted to~generalized linear models is presented by \emph{Delete or Merge Regressors} algorithm (\emph{DMR4glm}, \citealp{Proch}). It directly uses the agglomerative hierarchical clustering \citep{clustering} to build a hierarchical structure of groups that are being compared. 
Experimental studies \citep{Proch} show that \emph{Delete or Merge Regressors}'s performance is better than \emph{CAS-ANOVA}'s when it comes to the accuracy of the resulting model. The \emph{Delete or Merge Regressors} (DMR) method was first implemented in the \pkg{DMR} \proglang{R}~package \citep{DMR} and is reimplemented for a broader number of model families in the \factorMerger package.

The approach presented in this article extends approaches presented above in following ways:
\begin{itemize}
\item in comparison to pairwise tests, the Merging Path Plot is easier to interpret,
\item the \factorMerger visualizations are created based on 
\pkg{ggplot2} \citep{ggplot2} package and are easy to customize,
\item in comparison to the Fused LASSO, the Merging Path Plot is based on the likelihood ratio test statistic, which has known asymptotic properties. This allows us to calculate p-values for two nested models,
\item in comparison to the Fused LASSO, the obtained group estimates are not biased and are easier to interpret,
\item in comparison to the {DMR}, the Merging Path Plot can be extended to a wider variety of regression models, like generalized linear models and survival regression models,
\item as we will show later, in comparison to {DMR}, the resulting structure of groups in the Merging Path Plot is more stable. 
\end{itemize}

In the next section we present the methodology beyond the \factorMerger package.

\section{The Merging Path Plot}
\label{sec:meth}

The \pkg{factorMerger} package fits series of nested models. Each consecutive model is created based on the fusion of two closest groups with respect to the LRT-based distance. The hierarchy of obtained models along with distances between them are summarized in a consistent graphical way. Below we describe this procedure in a more formal way.

Let $k$ stand for the number of groups, while $n_i$ stands for the number of observations in group $i \in \{1, ..., k\}$.
Let $y_{ij}$ denote an observed value of the variable of interest for observation $j \in \{1, ..., n_i\}$ in group $i$. We assume that $y_{ij} \sim F(\theta_i)$, where $F$ is a distribution from exponential family parametrized by $\theta \in \Theta$.

The global null hypothesis is 
$$
H_0: \forall_{i \in \{1, ..., k\}} \theta_i = \theta_1
$$
and can be tested with the Likelihood Ratio Test for $k$ samples. If the global null hypothesis is rejected, then in the post-hoc analysis we are looking for groups with equal distributions, that is sets of indexes $J$ such as 
$\forall_{i,j \in J} \theta_i = \theta_j$

In the Merging Path Plot these sets are obtained in an iterative fashion. In every step two groups are merged into a single one. This step is repeated as long as there is more than one group.
The general sketch of~the~algorithm is described below.

The merging procedure begins with a full model --- with all original groups --- and iteratively merges merges pairs of groups until all of them are combined. 
For considered families of distributions, we use generalized linear models or Cox proportional hazard model. Each merging of two groups reduces by one the number of degrees of freedom of a model. In a single iteration pairs \emph{worth~fusing} are considered and the pair which optimizes an objective function is merged. 
In general, any model statistic may be used as an objective function, but here we are using the likelihood statistic. We specify it in details in the next section. A general formulation of the merging procedure is described in Algorithm 1.

\begin{algorithm}[H]
\caption{\label{algorithm1}The outline of the Merging Path Plot algorithm implemented in \factorMerger}
\label{outline}
\begin{algorithmic}[2]

\Function{MergeFactors}{$responseVariable, groupingVariable,
adjacent$}
\State{$currentModel:= createModel(responseVariable, groupingVariable)$}
\State{$mergingPath := list(currentModel)$}
\While{$|levels(groupingVariable)| \geq 1$}
\State{$pairsSet := generatePairs(groupingVariable, responseVariable, adjacent$)}
    \State{$selectedPair :=\mathrm{argmax_{pair \in pairsSet}}objectiveFunction
    (pair, responseVariable,\linebreak groupingVariable)$}    
    \State{$groupingVariable := mergeLevels(groupingVariable, selectedPair)$}
    \State{$currentModel := createModel(responseVariable, groupingVariable)$}
   \State{$mergingPath := add(mergingPath, currentModel) $}
 \EndWhile
 \State{return(mergingPath)}
    \EndFunction
\end{algorithmic}
\end{algorithm}

\begin{figure}[h!tbp]
\centering
\includegraphics[width=\textwidth]{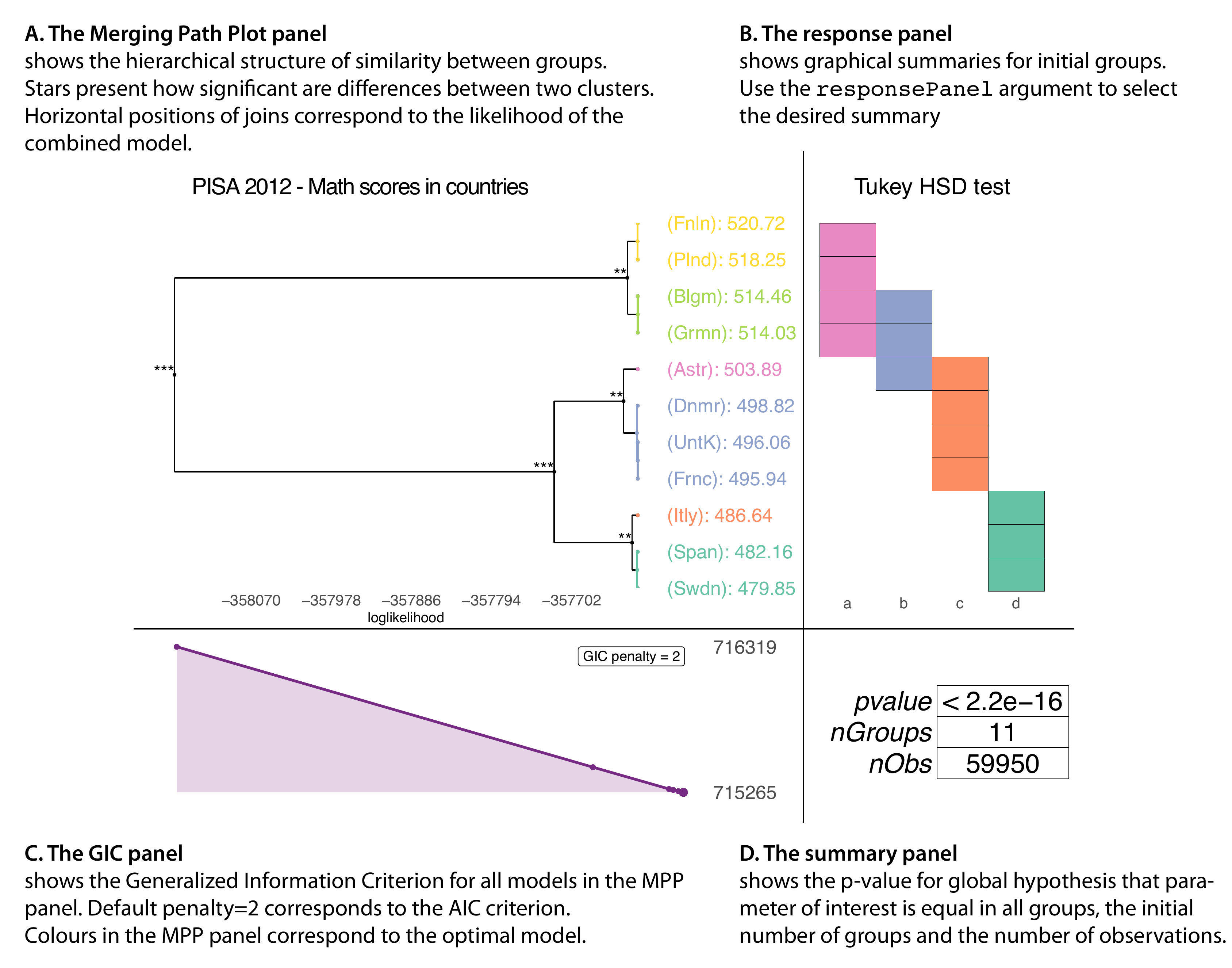}
\caption{\label{fig:PISAtree}
Four panels of the \pkg{factorMerger}'s visualization for the PISA dataset for 11~countries. Panel A summarizes the structure of group similarities. It shows the list of  models returned in the Algorithm 1. The OX axis presents values of the log-likelihood function for each model from the list. Labels on the right margin present averages of variable of interest for different groups. Stars placed in different joins of the tree summarize pairwise tests for selected groups of variables. Panel B summarizes the distribution of variable of interest in each group. The summary plotted in this panel may be changed depending on the model family. Panel C shows the Generalized Information Criteria for each model from the list. Panel D presents results from the test for the global null hypothesis. Colors in panels A~and~B~are consistent and correspond to an optimal segmentation of groups based on the GIC score.}
\end{figure}

The result of the Algorithm 1 is a list of $k$ shrinking models $\mathcal{M}_i$, where $i \in \{1, ..., k\}$. In~\factorMerger these models are presented in a graphical way in a \emph{Merging Path Plot} along with diagnostic criteria like Generalized Information Criteria and other graphical summaries. 
The Merging Path Plot contains four panels that encapsulate all important information in a~compact form.
An example of these panels is presented in Figure \ref{fig:PISAtree}.

The statistics presented in this plot are described in the following subsections.

\subsection{Model families}

The Merging Path Plot algorithm can be performed for any likelihood-based model. Current version of the \factorMerger package supports following parametric models: 

\begin{itemize}
\item one-dimensional Gaussian (with the argument \code{family = "gaussian"}).  Here 
$$
y_{ij} \sim \mathcal{N}\left(\mu_{j}, \sigma^2\right)
$$
and corresponding logarithm of likelihood

$$
l\left(\mu, \sigma | y\right) = 
-\frac{n}{2} \log{\left(2\pi\right)} -\frac{n}{2} \log{\left(\sigma^2\right)} -\sum_{j=1}^k\sum_{i=1}^{n_j}\frac{1}{2}\left(y_{ij} - \mu_j\right)^2/ \sigma^2.
$$

Group summaries are averages -- maximum likelihood estimates for $\mu_j$.

\item n-dimensional Gaussian (with the argument \code{family = "gaussian"}). Here $Y_{ij}$ and $M_j$ are vectors and

$$Y_{ij} \sim \mathcal N \left(M_j, \Sigma\right).$$

The corresponding logarithm of likelihood function

$$
l\left(M, \Sigma | Y\right) = 
-\frac{n}{2} \log{\left(2\pi\right)} -\frac{n}{2} \log{\left(|\Sigma|\right)} -
\sum_{j=1}^k\sum_{i=1}^{n_j}\frac{1}{2}\left(Y_{ij} - M_j\right)^T\Sigma^{-1}\left(Y_{ij} - M_j\right).
$$

Note that both one-dimensional and n-dimensional Gaussian models use \code{family = "gaussian"}. However, the visual summary of n-dimensional data requires additional preprocessing -- dimensionality reduction, and thus, it is considered a~separate category.
Group summaries are averages.

\item binomial (with the argument \code{family = "binomial"}). Here
$$y_{ij} \sim \mathcal{B}\left(p_{j}, 1\right).$$
After adding the logit link function 
$$\log\left(\frac{p_j}{1 - p_j}\right) = \beta_j$$
one may write the logarithm of likelihood


$$l\left(\beta|y\right) = \sum_{j=1}^k\sum_{i=1}^{n_j} y_{ij}\beta_j  - y_{ij} \log \left(1 + \exp \beta_j\right) + (1 - y_{ij}) \log\left(1+\exp \beta_j\right).$$

Group summaries are proportions of successes as estimates of $p$.

\item survival (with the argument \code{family = "survival"}). Here we consider the \emph{Cox proportional hazard model} \citep{coxph}.
Let $\lambda_0(t)$ be the baseline hazard function, where $t$~denotes time. Then the hazard function for group $j$ may be expressed as

$$
\lambda_j(t) = 
\lambda_0(t)\cdot \exp(\alpha_j).
$$

Corresponding logarithm of partial likelihood is

$$
l\left(\alpha\right|y) = 
	\sum_{i,j:C_{ij} = 1} 
		\left(
			\alpha_j -
			\log{
				\left(
				\sum_{kl:y_{kl}\geq y_{ij}} 
				\exp{\left(\alpha_k\right)}		
				\right)
			}
		\right).
$$
where $C_{ij}$ is the censoring status, $C_{ij} = 1$ what means that the observation $i$ from group $j$ is not censored.
For this model hazard ratios are the group summaries.

\end{itemize}

The fusing algorithm used in The Merging Path Plot is based on the Likelihood Ratio Test statistic defined as

\begin{equation} \label{LRTstat}
LRT(\M_1;\M_2) = 2 \cdot l (\widehat{\beta_{\M_2}}|y) - 2 \cdot l (\widehat{\beta_{\M_1}}|y),
\end{equation}
where $\mathcal{M}_1$ and $\mathcal{M}_2$ are two nested models. Each model corresponds to a grouping of observations. Groupings for both models are equal except that two groups in $\M_2$ are merged in one group in $\M_1$.
The higher the $LRT(\M_1;\M_2)$, the more different are the merged groups. One may interpret the $LRT(\M_1;\M_2)$ as a distance between groups for model $\M_1$~and~$\M_2$. 

The advantage of the \emph{LRT statistic} is the known asymptotic behavior (see \citealp{wilks1938large}). For nested models $\M_2$ and $\M_1$ that differ by one degree of freedom it holds
$$
LRT(\M_1;\M_2) \overset{n \rightarrow \infty}{\sim} \chi^2_1.
$$
This asymptotic distribution is used in \factorMerger to present statistical significance of group joins with the argument \code{panelGrid = TRUE} of the \fct{plot.factorMerger} function. 

\begin{figure}[H]
\centering
\includegraphics[width=0.8\textwidth]{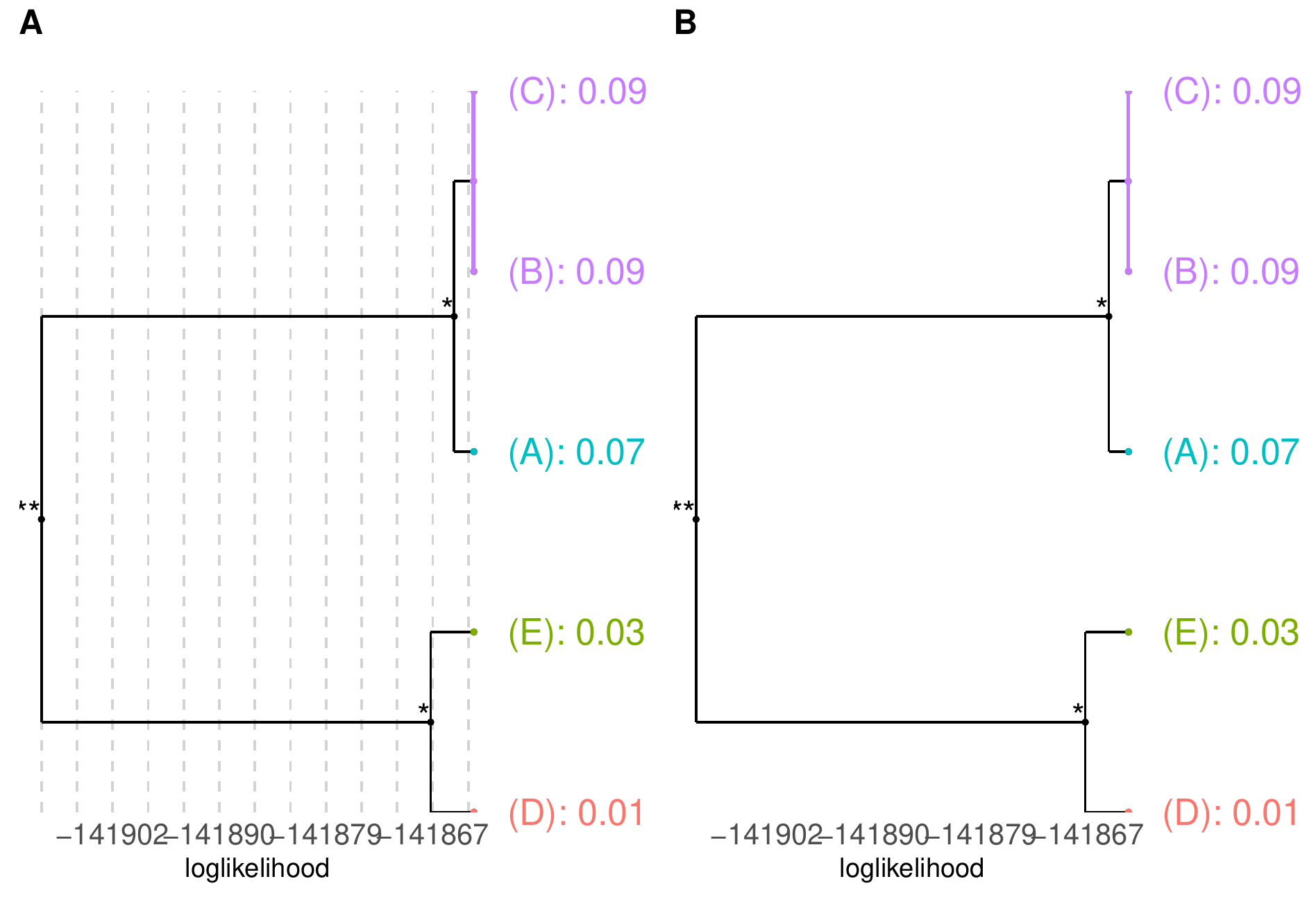} \caption{The Merging Path Plot with panel grid (Panel A) and without panel grid (Panel B). In Panel~A~each interval in the OX axis corresponds to the 0.95 quantile of chi-square distribution with one degree of freedom. Models distant more than by the length of this interval may be considered significantly different.
}
\end{figure}

\subsection{Group summaries}

The right panel of the visualization shows graphical summaries of the variable of interest in groups. Use the \texttt{responsePanel} argument to choose how groups shall be presented.

\begin{figure}[h!tbp]
\centering
\includegraphics[width=\textwidth]{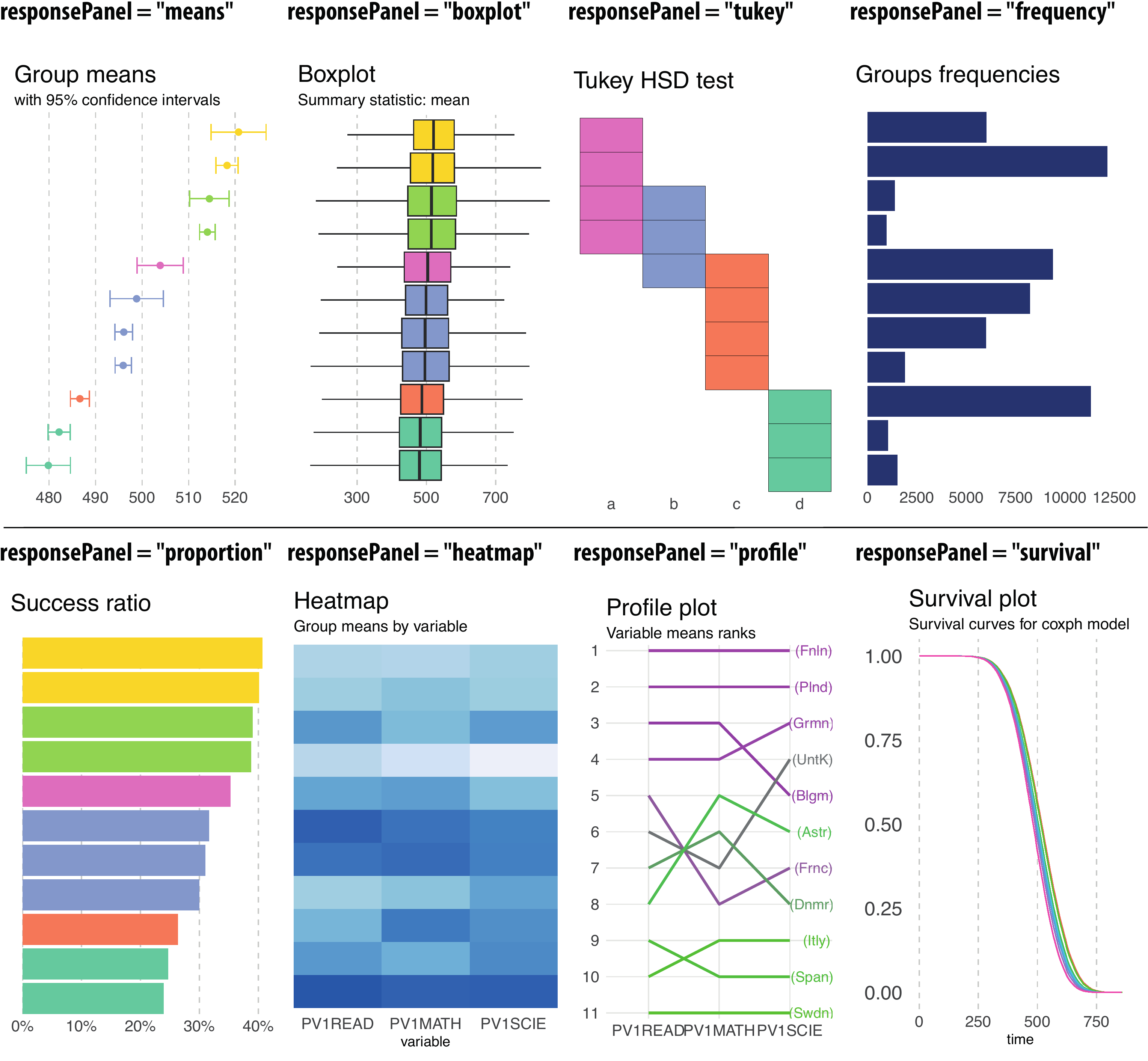}
\caption{\label{fig:panele}Available options for the \texttt{responsePanel} argument. Different panels are designed to highlight different summaries of groups.}
\centering
\includegraphics[width=0.75\textwidth]{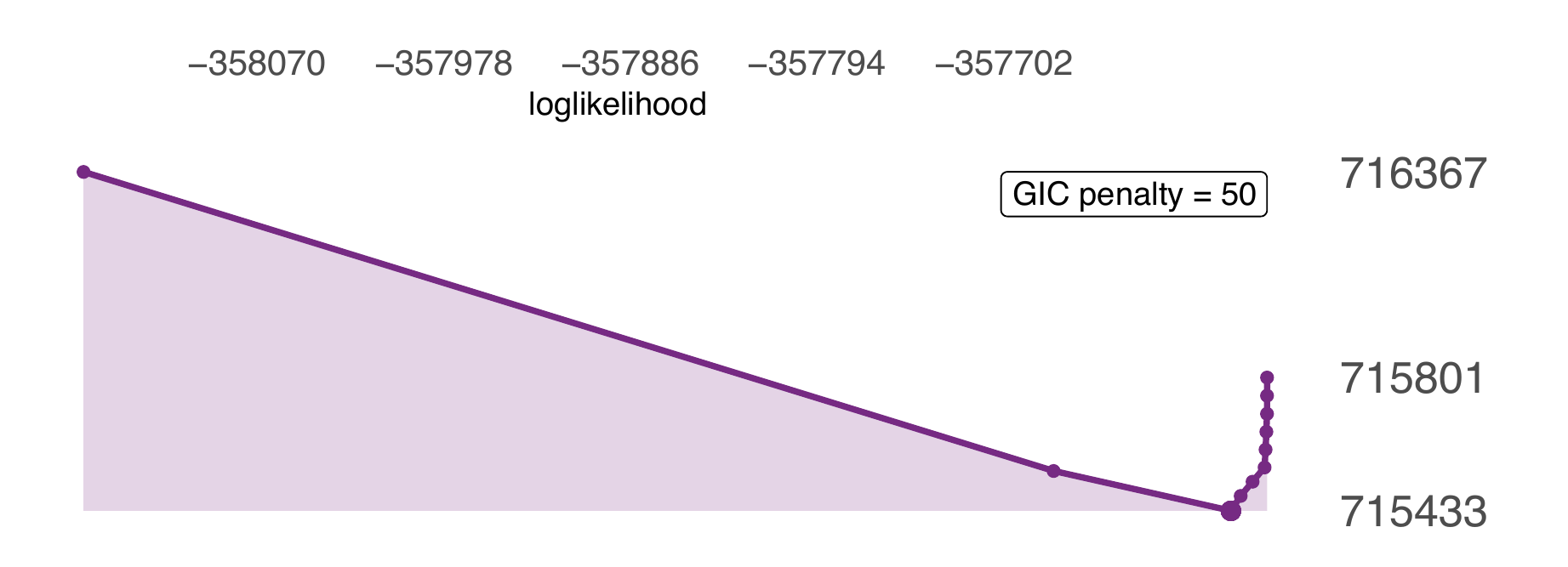}
\caption{\label{fig:FM0}
The GIC plot. The OX axis corresponds to the log-likelihood for a model. The OY axis corresponds to the GIC score for a model. Each dot denotes a single model from the merging path. GIC scores for the best, smallest and largest models are presented in the right axis.}
\end{figure}

Available options are shown in Figure \ref{fig:panele}. Depending on the family of the variable of interest, different summaries are appropriate. Possible combinations are listed in Table~\ref{tab:responsePanelT}.

\begin{table}[H]
\centering 
\begin{tabular}[t]{c|c|c|c}
 \textbf{} & \textbf{} & \textbf{family} & \textbf{} \\
 \textbf{responsePanel} & \textbf{gaussian} & \textbf{binomial} & \textbf{survival} \\
\hline frequency & + & + & + \\
\hline means & + & &  \\
\hline boxplot & + & &  \\
\hline tukey & + & &  \\
\hline heatmap & + & &  \\
\hline profile & + & &  \\
\hline proportion & & + &  \\
\hline survival & & & + \\
\hline 
\end{tabular}
\caption{\label{tab:responsePanelT}Different types of the graphical summary are appropriate for different model families. Pluses denote which \texttt{responsePanel} may be used for which model family. Examples for each type of panel are presented in Figure \ref{fig:panele}.}
\end{table}

\subsection{Optimal grouping selection}

The Merging Path Plot algorithm returns a~collection of models of different sizes / different numbers of groups. In order to select the best model, the optimization criterion must be specified in the first place.
There are three metrics available in the \pkg{factorMerger}:
\begin{itemize}
\item \emph{Generalized Information Criterion} with an additional penalty parameter. If this option is selected, the model with the lowest GIC is returned.
\item \emph{p-value} for the Likelihood Ratio Test against the full model. If we go with this metric, we choose the latest model in the merging path whose p-value for the LRT test against the full model is greater than a given threshold.
\item \emph{log-likelihood} of a model. A similar search is performed as in the previous point, but with models' log-likelihood as the model score.
\end{itemize}

The most natural approach is to pick a~model that minimizes the Generalized Information Criteria
$$
GIC(\M) = -2 l(\M) + p |\M|.
$$
Here $|\M|$ denotes the number of groups in model $\M$, while $p$ is a penalty for model complexity. GIC corresponds to Akaike Information Criterion (AIC) for $p=2$ or Bayesian Information Criterion (BIC) for $p=\log(n)$, where $n$ is the number of observation.

To ease the selection of the best model, the bottom-left panel presents GIC scores for models in the merging path in the GIC plot. An example of such plot is presented in Figure \ref{fig:FM0}.

\subsection{The Fusing Strategy}

The Algorithm \ref{algorithm1} presents a general strategy for merging groups. The fully adaptive strategy is time-consuming and may be slow for a large number of groups. Thus, in the \factorMerger package we have implemented four versions of the merging algorithms. These versions are summarized below.

Depending on the specific goal, some steps of the Algorithm 1 may be performed differently. Possible options are:
\begin{itemize}
\item \texttt{method = {"}adaptive{"}}. The objective function is the logarithm of likelihood. The set $pairsSet$ contains all possible pairs of groups available in a given step. Pairwise LRT distances are recalculated at every step. This option is the slowest one since it requires the largest number of comparisons. It requires $\text{O}(k^3)$ model evaluations. 
\item \texttt{method = {"}fast-adaptive{"}}. 
Note that computing an objective function can be expensive and, especially for big datasets, it may be beneficial to limit the set of pairs that shall be compared. Also note that it is more likely that a~pair~of~levels $i$ and $j$ is selected to merge if~corresponding group averages are close. In this option, the objective function is the logarithm of likelihood, but the $pairsSet$ is generated differently in the following way: 
for Gaussian family of response, at the very beginning, the groups are ordered according to increasing averages and consequently $pairsSet$ contains only pairs of closest groups. 
For other families, the order corresponds to beta coefficients in a regression model. The detailed rules of ordering levels are given in Table~\ref{tab:ordering}.
This option is much faster than \texttt{method = {"}adaptive{"}} and requires $\text{O}(k^2)$ model evaluations.
\item \texttt{method = {"}fixed{"}}. This option is based on the DMR algorithm introduced in \cite{Proch}. It was extended to cover survival models (however, for survival models there are no theorems of model selection consistency yet proven). The largest difference between this option and the \texttt{method = {"}adaptive{"}} is that in the first step a pairwise distances are calculated between each pair of groups based on the $LRT$ statistic. Then the agglomerative clustering algorithm is used to merge consecutive pairs. It means that pairwise model differences are not recalculated as LRT statistics in every step but the \texttt{complete linkage} is used instead. 
This option is very fast and requires $\text{O}(k^2)$ comparisons.

\item \texttt{method = {"}fast-fixed{"}}. This option may be considered as a~modification of~the \texttt{method = {"}fixed{"}}. Here, similarly as in the \texttt{{"}fast-adaptive{"}} version, we assume that if groups $A, B$ and $C$ are sorted according to their increasing beta coefficients, then it is worthwhile to join groups $A$ and $B$ or groups $B$ and $C$ (but not groups $A$ and $C$). This assumption enables implementation of the \emph{complete linkage} clustering more efficiently and in a mode dynamic manner. The biggest difference is that in the first step we do not calculate the whole matrix of pairwise differences, but instead only differences between consecutive groups are measured. Then in each step only a~single distance is calculated. This reduces the number of model evaluations to $\text{O}(k)$. A~detailed description of beta coefficients is given in Table \ref{tab:ordering}.
\end{itemize}

\begin{figure}[h!tbp]
\centering
\includegraphics[width=0.9\textwidth]{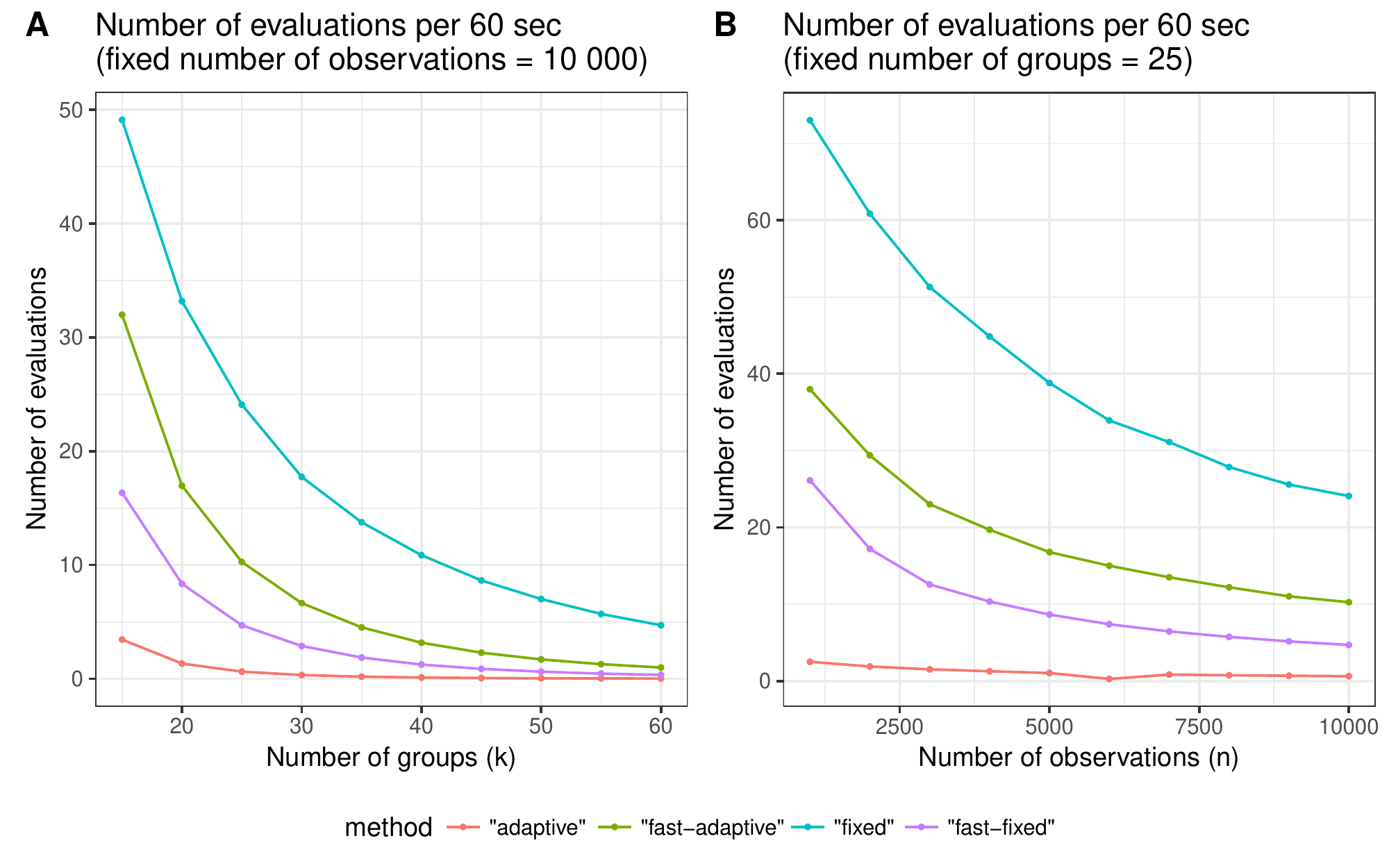}
\caption{
\label{fig:nEvalsPerMinute}
Average number of evaluations per 60 seconds for different number of groups (Panel A) and different sample sizes (Panel B). Fastest algorithms are those which limit comparisons only to consecutive groups (\code{"fast-adaptive"} and \code{"fast-fixed"}).
For 10 observations methods \code{"fast-fixed"} and \code{"fast-adaptive"} are 5 times faster than method \code{"adaptive"}. For 60 observations those evaluation time ratios grow up to 200 and 42, respectively for \code{"fast-fixed"} and \code{"fast-adaptive"} against \code{"adaptive"}.
Use following \pkg{archivist} (\cite{archivist}) links \texttt{aread("MI2DataLab/factorMerger/materials/7019f")} and \texttt{aread("MI2DataLab/factorMerger/materials/7d1c9")} to access detailed results.}
\centering
\includegraphics[width=0.7\textwidth]{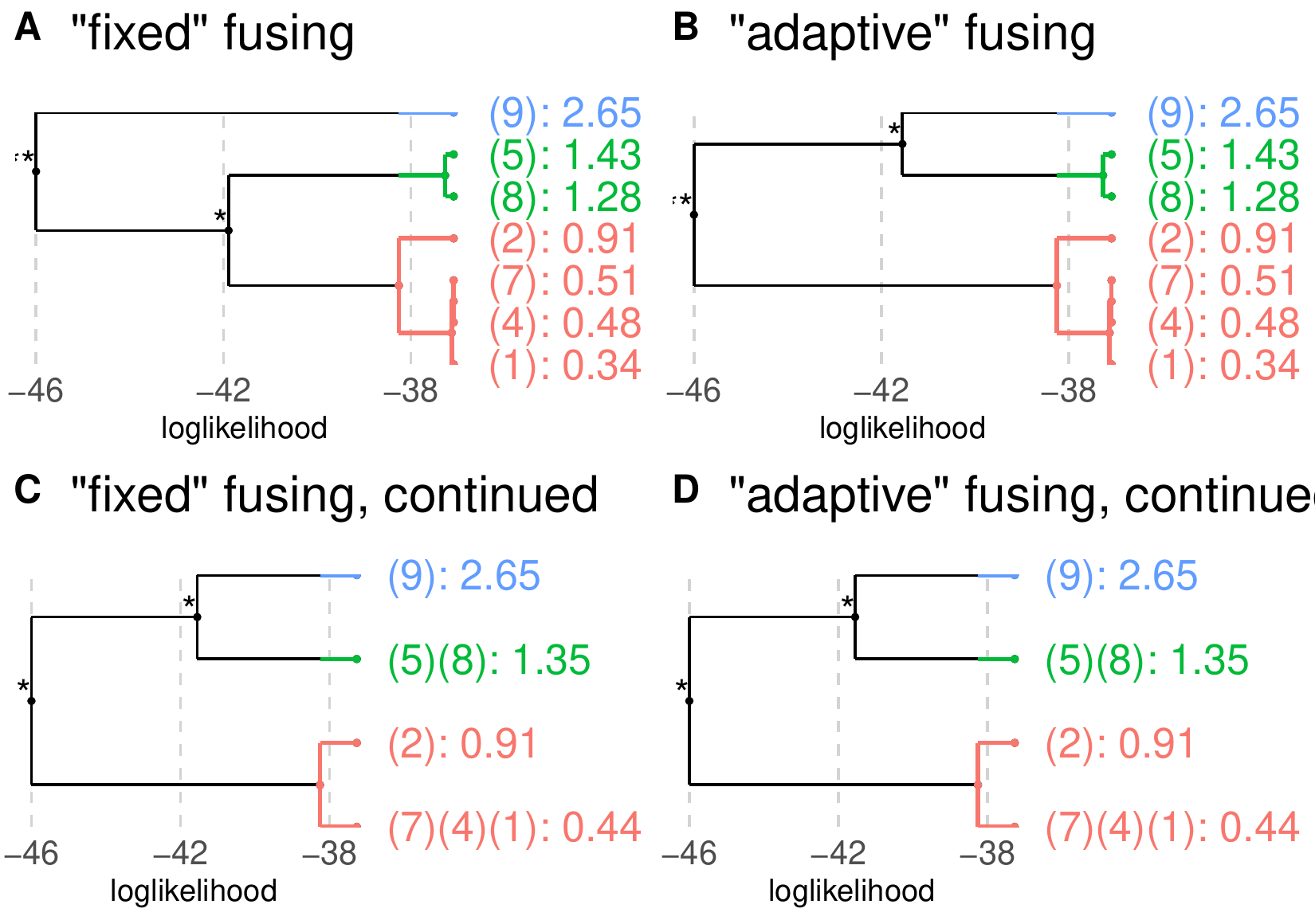}
\caption{
\label{fig:differentTrees}
Comparison of methods \code{"fixed"} (left panels) and \code{"adaptive"} (right panels).
We start with a sample consisting of 7 subgroups (top panels). First four steps of both algorithms are the same, but then the \code{"fixed"} algorithm chooses to merge groups (8)(5) and (9), while the \code{"adaptive"} algorithm goes with groups (1)(4)(7)(2) and (8)(5). The latter results in a~model with higher log-likelihood (\code{"fixed"}: -41.89 vs. \code{"adaptive"}: -41.57). 
Note that if we choose different starting point (bottom panels), the \code{"fixed"} algorithm changes its path.
}
\end{figure}

Described options differ in two ways. First, they differ in terms of computational time. The fastest option is to preliminarily sort groups and then use the dynamic complete-linkage hierarchical algorithm which allows for joining only adjacent groups. The slowest option is to calculate pairwise differences between groups after each fusion. Time performance comparisons are presented in Figure \ref{fig:nEvalsPerMinute}.

\begin{table}[h!tb]
\centering \begin{tabular}[t]{c|p{9cm}}
\hline \textbf{family} & \textbf{ordering statistic for a given group} \\
\hline one-dimensional Gaussian & average in a group \\
\hline multi-dimensional Gaussian & average in a group after the Kruskal's non-metric multidimensional scaling \citep{MASS} to a one-dimensional space \\
\hline binomial & proportion of successes in a group\\
\hline survival & logarithm of a hazard ratio for a group\\
\hline 
\end{tabular}
\caption{\label{tab:ordering}Factor ordering by model family for \texttt{method = {"}fast-adaptive{"}} and \texttt{method = {"}fast-fixed{"}}}
\end{table}

At the same time, the slowest option is the most accurate one in terms that it gives models paths with the highest log-likelihood and ensures stability. A simple example that brings closer those characteristics is visualized in Figure \ref{fig:differentTrees}.

\section{Examples}
\label{sec:examples}

The \pkg{factorMerger} package is highly customizable.
In this section we present three different case studies to illustrate the use of \factorMerger in real-world examples. Each scenario is associated with a particular model family. It also presents specific function arguments in action. 

\subsection{Academic performance in mathematics of 15-year-old kids around the world}
\label{subsec:gauss}

Programme for International Student Assessment (\citealp{pisa2012}) is a study maintained by \textit{Organisation for Economic Co-operation and Development} (OECD) to gather information on students' academic performance in mathematics, science and reading. The performance is expressed by the \emph{plausible values} that are normalized to have conditionally Gaussian distribution.

The \pkg{factorMerger} package provides a student-level dataset \code{pisa2012} for 271322 students from  43 countries with tree plausible values together with the country affiliations. 
The data is a~weighted version of the original data. Find more in \cite{pisa2012lite}. 

Following instructions create the Merging Path Plot for differences between countries concerning performance in  mathematics.

\begin{CodeChunk}
\begin{CodeInput}
R> library("factorMerger")
R> library("dplyr")
R> data("pisa2012")
R> oneDimPisa <- mergeFactors(response = pisa2012$math,
+    factor = pisa2012$country, method = "fast-fixed")
\end{CodeInput}
\end{CodeChunk}

Note that only one command is needed to perform the merging procedure. To speed up the evaluation, we use \code{"fast-fixed"} method.

We can use the obtained object to display the history of merging --- each row of the table describes one step of the algorithm. 

\begin{CodeChunk}
\begin{CodeInput}
R> mergingHistory(oneDimPisa, showStats = TRUE) %>% 
+     head(5)
\end{CodeInput}
\end{CodeChunk}

\clearpage

\begin{CodeChunk}
\begin{CodeOutput}
   groupA groupB    model pvalVsFull pvalVsPrevious
 0               -1606256     1.0000         1.0000
 1 (Swdn) (SlvR) -1606256     0.9932         0.9932
 2 (RssF) (Span) -1606256     0.9997         0.9805
 3 (Chil) (Mlys) -1606256     0.9999         0.9459
 4 (Frnc) (UntK) -1606256     1.0000         0.9213
\end{CodeOutput}
\end{CodeChunk}

Here, in the second step, Russian Federation \code{(RssF)} and Spain \code{(Span)} were joined. Log-likelihood, whose value is given in the \code{model} column, decreased marginally, p-values for the LRT test against the full model and against the previous model were 0.9997 and~0.9805, respectively. This means that the data partition created after two joins is equally good as the the previous and initial partitions. 

In order to create the optimal data partition, the model with the lowest AIC in the merging path (GIC penalty = 2) is chosen. Final grouping names are concatenations of original levels' names.

\begin{CodeChunk}
\begin{CodeInput}
R> aicPrediction <- cutTree(oneDimPisa, stat = "GIC", value = 2)
R> aicPrediction %>% table() %>% head(4)
\end{CodeInput}
\begin{CodeOutput}
  (Clmb)       (Brzl) (Mntn)(Urgy) (Chil)(Mlys)
   8902        38525          763        10872
\end{CodeOutput}
\end{CodeChunk}

We can also see the final data partition in a table. Below original group labels are printed in abbreviated form (\code{orig}) together with their final cluster name (\code{pred}). For example, Poland \code{(Plnd)} ends up in the \code{(Cand)(Plnd)} group, which consists of two members: Poland and Canada \code{(Cand)}. 

\begin{CodeChunk}
\begin{CodeInput}
R> getOptimalPartitionDf(oneDimPisa,  stat = "GIC", value = 2) %>% head(5)
\end{CodeInput}
\begin{CodeOutput}
     orig         pred
 1 (RssF) (RssF)(Span)
 2 (Blgm) (Grmn)(Blgm)
 3 (Grmn) (Grmn)(Blgm)
 4 (Kore)       (Kore)
 5 (Plnd) (Cand)(Plnd)
\end{CodeOutput}
\end{CodeChunk}

\subsection{Happiness in Europe}
\label{subsec:binom}

This section uses the \code{ess} dataset included in the \pkg{factorMerger} package.  
The data concerning happiness of~21~European countries is based on the European Social Survey (ESS) \citep{ess}. A~binary variable called \emph{happy} specifies if a given individual considers himself or herself a happy person (or, more precisely, whether his/her answer to the question "Taking all things together, how happy would you say you are?" was greater than 5). The data is~weighted according to the original weights given by ESS. A total number of rows in \code{ess} is~200 075; there are 21 countries included.  

By default the \fct{plot.factorMerger} function uses GIC with the penalty equal to 2 (i.e. Akaike Information Criterion).

\begin{figure}[h!bt]\centering

\includegraphics[height=0.8\textheight]{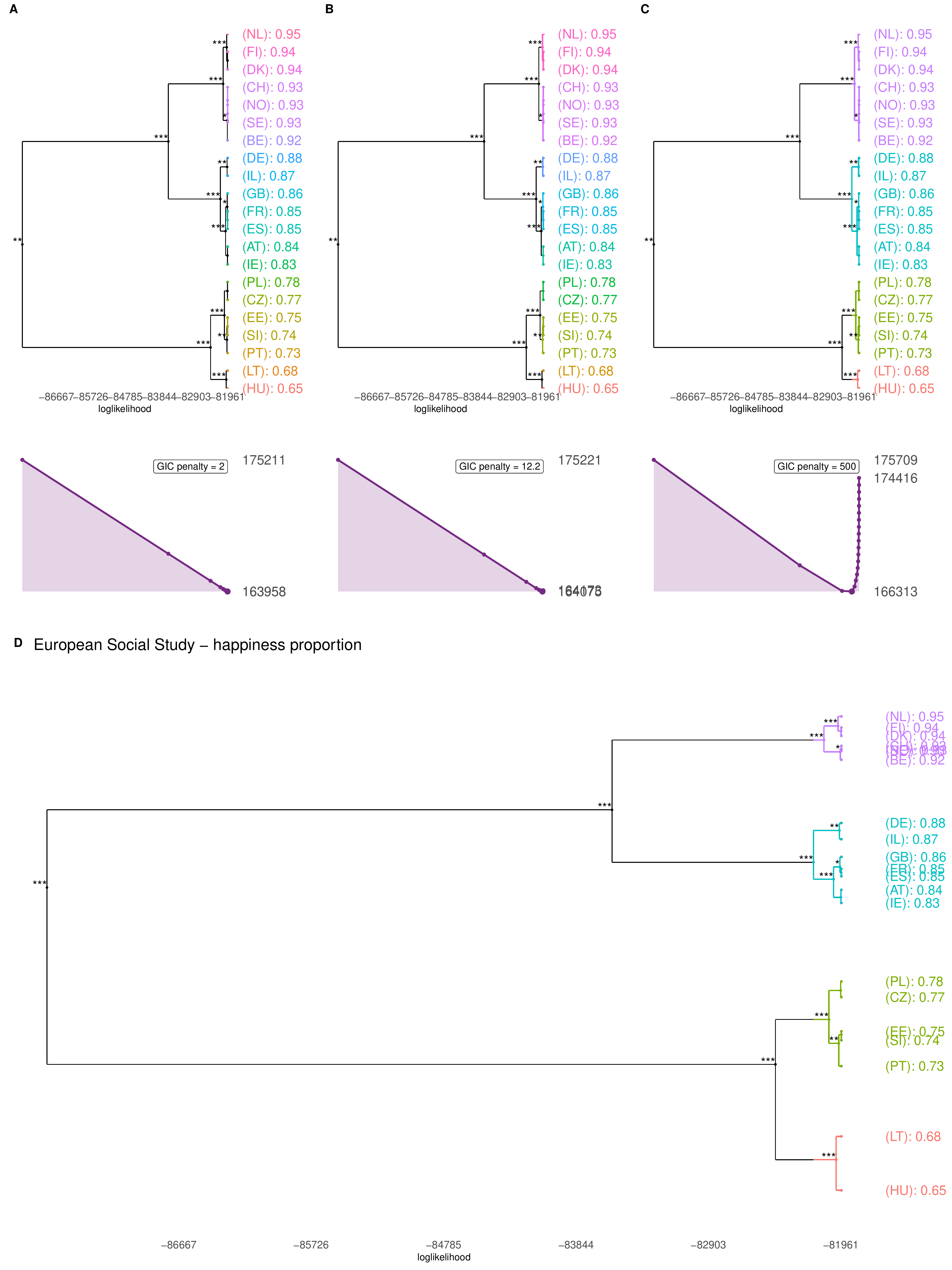}

\caption{Results for GIC with different penalties: AIC with penalty = 2 (Panel A), BIC with penalty = 12.2 (Panel B) and GIC with penalty = 500. Optimal numbers of groups for those penalties are: 17, 9, 4. 
The Merging Path Plot for GIC with penalty = 500 is presented in Panel D. 
Positions of nodes on the OY axis correspond to fractions of happy citizens in a given country / group of countries.}
\end{figure}

\clearpage

\begin{CodeChunk}
\begin{CodeInput}
R> library("factorMerger")
R> data("ess")
R> happyMerge <- mergeFactors(ess$happy, ess$country,
+     family = "binomial", method = "fast-fixed")
R> p1 <- plot(happyMerge, panel = "GIC", title = "", panelGrid = FALSE)
\end{CodeInput}
\end{CodeChunk}

However, in some cases it may not be restrictive enough. Since the number of observation is large, we may use a much larger penalty.

\begin{CodeChunk}
\begin{CodeInput}
R> p3 <- plot(happyMerge, panel = "GIC", penalty = 500, title = "", 
+     panelGrid = FALSE)
\end{CodeInput}
\end{CodeChunk}

\subsection{Survival of cancer patients}
\label{subsec:surv}

In this example we use data from The Cancer Genome Atlas Project \citep{tcga} from the \pkg{RTCGA.clinical} package \citep{RTCGA.clinical}. TCGA is a~public-funded project that aims to catalogue and discover major cancer-causing genomic mutations to create a comprehensive \emph{atlas} of cancer profiles. The \pkg{RTCGA.clinical} package provides a snapshot of this clinical data created on 2015-11-01. In our example we focus on patients who suffer from breast cancer and are treated with different drugs. We are interested whether drug treatments may be grouped according to their effectiveness.

The dataset \code{BRCA} used in this example is included in \pkg{factorMerger} package. First, some data preprocessing is performed.

\begin{CodeChunk}
\begin{CodeInput}
R> library("factorMerger")
R> library("dplyr")
R> library("forcats") 
R> library("survival") 
R> data("BRCA")
R> BRCA <- BRCA %>% filter(!is.na(drugName))
R> drugName <- fct_lump(BRCA$drugName, prop = 0.05) 
R> brcaSurv <- Surv(time = BRCA$time, event = BRCA$vitalStatus)
R> drugMerge <- mergeFactors(response = brcaSurv, factor = drugName,
+     family = "survival", method = "adaptive")
\end{CodeInput}
\end{CodeChunk}

Now we can plot the result. By default four panels are included in the plot, the tree is colored to denote final clusters, and nodes are spaced at equal distances. 

One may add some customization to the plot. In this example, the horizontal position of the nodes indicates the risk score of a group, a personalized title is added, and only two top panels are displayed (namely the tree plot and the survival plot). Moreover, nodes coloring corresponds to the survival plot's coloring, final clusters are visually specified with a use of a vertical line, and a custom palette is added. 

\begin{CodeChunk}
\begin{CodeInput}
R> plot(drugMerge, nodesSpacing = "effects", 
+     title = "BRCA: patient survival vs. drug treatment",
+     panel = "response", colorClusters = FALSE,
+     showSplit = TRUE, palette = "Dark2")
\end{CodeInput}
\end{CodeChunk}

\begin{figure}[H]\centering
\includegraphics[width=\textwidth]{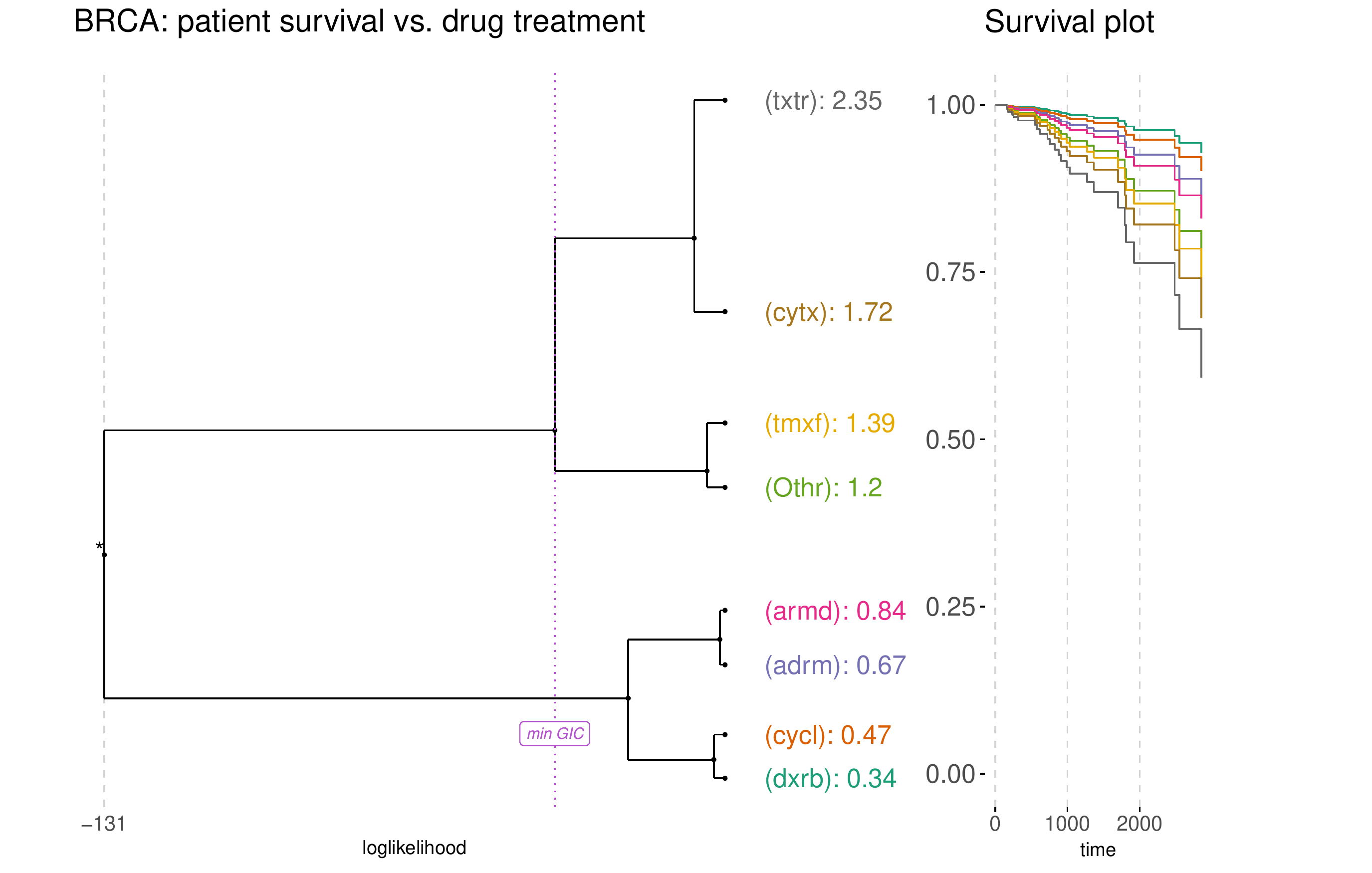}
\caption{
A customized \fct{plot.factorMerger} output. Colors of the OY axis' labels are guides for the right panel. Tree nodes are spaced according to group effects. A vertical line is added to mark the optimal data partition.
}
\end{figure}

\section{Summary and Future Directions}
\label{sec:summ}

The Merging Path Plot is a novel approach summarizing groups dissimilarities based on the LRT statistic. It is a useful tool to explore group similarities in k-sample comparisons.

In this article we have presented the methodology and its implementation. Examples presented in this article are limited to models with one independent variable, but the package \pkg{factorMerger} works also for models with weights or covariates.

The natural direction for future work is to extend this methodology to different classes of models. Instead of the Likelihood Ratio Test, other tests may be used. For example, the Wilcoxon test may be used for semi-parametric modeling. 

\section*{Computational details}
\label{sec:comp}
The results in this paper were obtained using
\proglang{R}~3.4.2 with the
\factorMerger~0.3.2 package. \proglang{R}~itself
and all packages used for the needs of this paper are available from the Comprehensive
\proglang{R} Archive Network (CRAN) at
\url{https://CRAN.R-project.org/}.

\section*{Acknowledgements}

We acknowledge the financial support from the \emph{NCN Opus grant 2016/21/B/ST6/02176}.

\bibliography{article_init}


\end{document}